\documentclass{article}

\usepackage{graphicx}       
\usepackage{subcaption}
\usepackage{float}
\usepackage{calc} 

\usepackage{amssymb}
\usepackage{amsmath}
\usepackage{mathpartir} 
\usepackage{datetime2}
\usepackage{booktabs}
\usepackage{makecell}
\usepackage[table]{xcolor}
\usepackage{array}
\usepackage{url}

\usepackage[margin=2.5cm]{geometry}

\definecolor{mygray}{gray}{0.6}

\newcommand{\comment}[1]{}

\begin{document}

\rule{\textwidth}{1pt}

\vspace{0.5cm}

\newcommand{\PaperTitle}[1]{%
  \begin{center}
    {\Large\bfseries #1\par}
  \end{center}
}



\newcommand{\supers}[1]{\textsuperscript{#1}}

\makeatletter

\newcommand{\PaperAuthors}[1]{\gdef\@paperauthors{#1}}
\newcommand{\PaperAffiliations}[1]{\gdef\@paperaffils{#1}}

\gdef\@paperauthors{}
\gdef\@paperaffils{}

\renewcommand{\maketitle}{%
  \begin{center}
    {\LARGE\bfseries \@title \par}
    \vspace{0.8em}
    {\large \@paperauthors \par}
    \vspace{0.8em}
    {\normalsize \@paperaffils \par}
  \end{center}
  \vspace{1.2em}
}

\makeatother

\title{SAMannot: A Memory-Efficient, Local, Open-source Framework for Interactive Video Instance Segmentation based on SAM2}

\PaperAuthors{%
Gergely Dinya\supers{1},
András Gelencsér\supers{1},
Krisztina Kupán\supers{2},
Clemens Küpper\supers{2},
Kristóf Karacs\supers{3},
Anna Gelencsér-Horváth\supers{1,3}\supers{*}%
}

\PaperAffiliations{%
\supers{1} Faculty of Informatics, Eötvös Loránd University, Budapest, Hungary\\
\supers{2} Max Planck Institute for Biological Intelligence, Seewiesen, Germany\\
\supers{3} Faculty of Information Technology and Bionics, Pázmány Péter Catholic University, Budapest, Hungary\\
\supers{*} Corresponding author: gha@itk.ppke.hu%
}
\maketitle







\section*{Abstract}
Current research workflows for precise video segmentation are often forced into a compromise between labor-intensive manual curation, costly commercial platforms, and/or privacy-compromising cloud-based services.
The demand for high-fidelity video instance segmentation in research is often hindered by the  bottleneck of manual annotation and the privacy concerns of cloud-based tools.
We present SAMannot, an open-source, local framework that integrates the Segment Anything Model 2 (SAM2) into a human-in-the-loop workflow.
To address the high resource requirements of foundation models, we modified the SAM2 dependency and implemented a
processing layer that minimizes computational overhead and maximizes throughput, ensuring a highly responsive user interface. 
Key features include persistent instance identity management, an automated ``lock-and-refine'' workflow with barrier frames, and a mask-skeletonization-based auto-prompting mechanism. 
SAMannot facilitates the generation of research-ready datasets in YOLO and PNG formats alongside structured interaction logs. Verified through animal behavior tracking use-cases and subsets of the LVOS and DAVIS benchmark datasets, the tool provides a scalable, private, and cost-effective alternative to commercial platforms for complex video annotation tasks.
\section*{Keywords}

instance segmentation; video annotation; instance tracking; instance labeling; 

\section*{Introduction}
The exponential growth of high-resolution video data across diverse sectors, including surveillance, human or animal behavioral analysis, autonomous driving, healthcare, agriculture and aerial imaging, has significantly outpaced the capacity for manual analysis, and
efficient methods are required to process large datasets by accelerating the annotation workflow.
The primary goal in many research and application settings is to detect one or more objects of interest in each frame, enabling the monitoring of their presence, the tracking of their motion, or to identify their behavior. 
Data annotation may be required to train a model (such as YOLO~\cite{yolo11_ultralytics} or DeepLabCut~\cite{dlc}) for automatic annotation, or to support human-in-the-loop workflows, where the aim is to reduce the required human attention and time as much as possible, since this is often a monotonous and low-skill task.
The need for efficient video instance annotation tool is reflected in the many frameworks available to support the processing and interpretation of large volumes of recorded data, as well as the extraction of domain-relevant information.

Established platforms such as Roboflow~\cite{roboflow} offer efficient annotation workflows, but require uploading data to an external platform, which is not possible, because it is unlawful or infeasible in many settings (e.g., medical data).
Proprietary tools such as Encord~\cite{encord} and Labelerr~\cite{labellerr} provide end-to-end video data labeling platforms with AI-assisted instance segmentation workflows (e.g., SAM~\cite{sam} integration), but they are commercial solutions: they are not open source and place the functionalities behind a paywall.
Although, CVAT~\cite{cvat} is offered both as a paid, hosted service and as a downloadable open-source platform, efficient instance tracking and propagation, integrated into the annotation workflow, are available only in the commercial (non-free) version.

There are also standalone open-source tools for manual image and video annotation that keep the data local however, they lack a seamless integration of modern foundation models, leaving a gap between purely manual tools and complex commercial platforms.
SILVI~\cite{kanbertay2025silvi} is an open-source video labeling tool aimed at spatio-temporally localized interaction annotation on top of object tracks, providing multi-view synchronization and streamlined track/ID correction and export for downstream model training. However, it supports only manual annotation and does not incorporate computer-vision-assisted labeling.
VGG~\cite{dutta2019vgg} runs locally, within a web-browser and minimal computational requirements. However, annotation is typically performed frame by frame, making the process relatively slow.

The development of the software was motivated by projects that continuously collect large volumes of video data --- often on the scale of hundreds of hours --- where reliable instance-level tracking and annotation are prerequisites for subsequent analysis. 
In such settings, development teams and even more so research groups often benefit from an open-access, customizable framework 
that 
remains easy to adopt even for non-technical users, and still provides an efficient back-end. 
A typical example arises in biological studies of animal behavior, where individuals of a group (e.g., each bird) must be tracked separately over long time periods of recordings. 
These data are usually captured under unconstrained conditions and therefore exhibit motion blur, frequent partial or long-term occlusions, and pronounced scale changes as subjects move toward or away from the camera; moreover, interactions between instances introduce additional ambiguities that further complicate instance-level labeling.
Consequently, annotation becomes both a major practical bottleneck—demanding fast, accurate workflows with minimal expert overhead—and a technically challenging task that mirrors the difficulties encountered by modern instance tracking and segmentation methods, even when the resulting data are not intended for release as a public benchmark.

Our contribution is an open-source general framework for instance-level segmentation that integrates SAM2~\cite{2024sam2} to enable efficient video instance segmentation and tracking in a human-in-the-loop setting while reducing the computation need of the original model.
We exploit the zero-shot generalization and prompt-guided mask generation for high-fidelity object segmentation of the SAM2 model. 
A prompt is a user-provided input (typically a point or a bounding box) that guides the model to identify and isolate a specific object.
We addressed two critical limitations to integrate SAM2 into a production-ready annotation workflow. 
First, we overcame the high memory overhead of long-sequence inference, we optimized the pipeline for commodity hardware.
This modification enables the processing of hundreds of frames on a single mid-range GPU while maintaining the low-latency response necessary for interactive use.
Second, SAM2 lacks an integrated system for persistent metadata and multi-object class management. To bridge this, we implemented a higher-level management layer that maps SAM2’s internal object slots to a persistent data structure. This layer enables the association of immutable instance IDs and user-defined semantic labels with each tracked mask, ensuring that metadata remains consistent during data export and across fragmented annotation sessions.
To support the annotation of long videos while maintaining a low memory footprint, we introduce an auto-prompting approach. This method re-initializes masks and inference states based on prior segmentation data.
Unlike existing solutions, SAMannot provides local, prompt-guided temporal propagation --- automatically extrapolating segmentation masks from annotated frames to others ---, ensuring that state-of-the-art AI assistance remains accessible without high-cost commercial subscriptions. 
The system is designed to run locally to 
ensure data privacy to support a broad range of users and application domains.

\section*{Implementation and architecture}

\subsection*{Overview}
\begin{figure}[H]
    \centering
    \includegraphics[width=\linewidth]{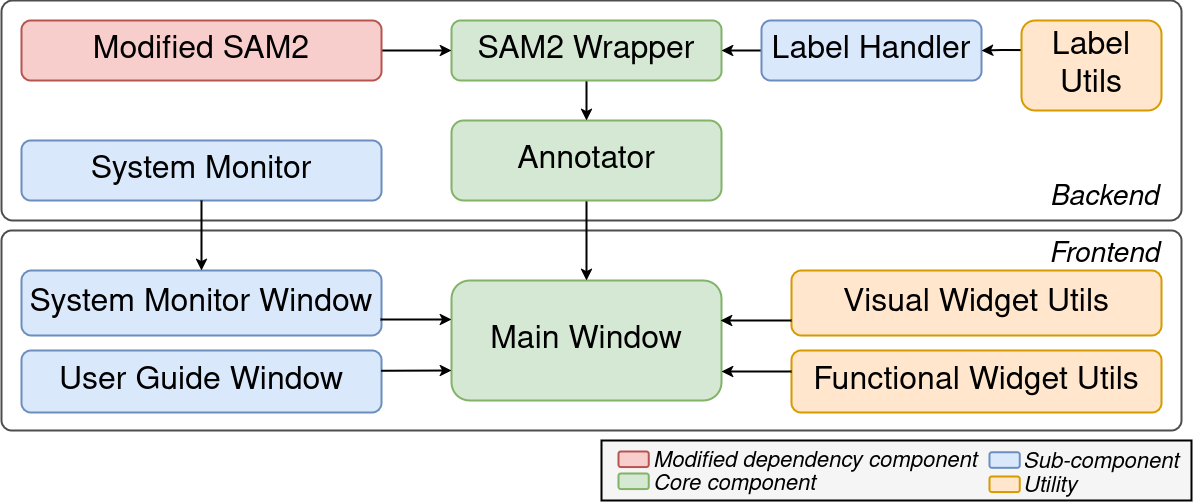}
    \caption{This figure provides a high-level overview of the software architecture, organized by module functionality. Arrows denote invocation direction. Green boxes indicate core pipeline components responsible for a broad range of tasks. Blue boxes represent subsidiary subsystems with well-defined roles within the pipeline. Yellow boxes correspond to utility classes, including data-structure components and auxiliary widgets extending Tkinter functionality. The red box denotes the underlying tracking dependency (SAM2).}
    \label{fig:SAMPL}
\end{figure}

The proposed SAMannot software consists of a Tkinter-based\cite{python_tkinter_docs} frontend, responsible for graphics and user interaction, and a backend that manages annotation logic and communicates with the SAM2 wrapper.
The main architecture of the software is visualized in Fig.~\ref{fig:SAMPL}

A dedicated System Monitor module independently tracks CPU and VRAM usage, providing real-time feedback through the System Monitor Window to prevent out-of-memory errors during long-range propagation.
The Graphical User Interface (GUI) logic is supported by two specialized utility modules that extend the standard Tkinter functionality. 
The GUI's visual and functional consistency is maintained through a set of specialized widget classes that inherit from and extend standard Tkinter components. 
Specifically, \emph{Visual Widget Utils} provides enhanced UI elements, such as a custom tooltip class and a circular progress bar widget, which allows for the aesthetic and real-time rendering of resource occupancy. 
Similarly, \emph{Functional Widget Utils} implements an extended listbox class that supports direct item editing via double-click events, enabling seamless interaction with labels and features without the need for auxiliary dialog windows.

\subsubsection*{Frontend}
\begin{figure}[H]
    \centering
    \includegraphics[width=\linewidth]{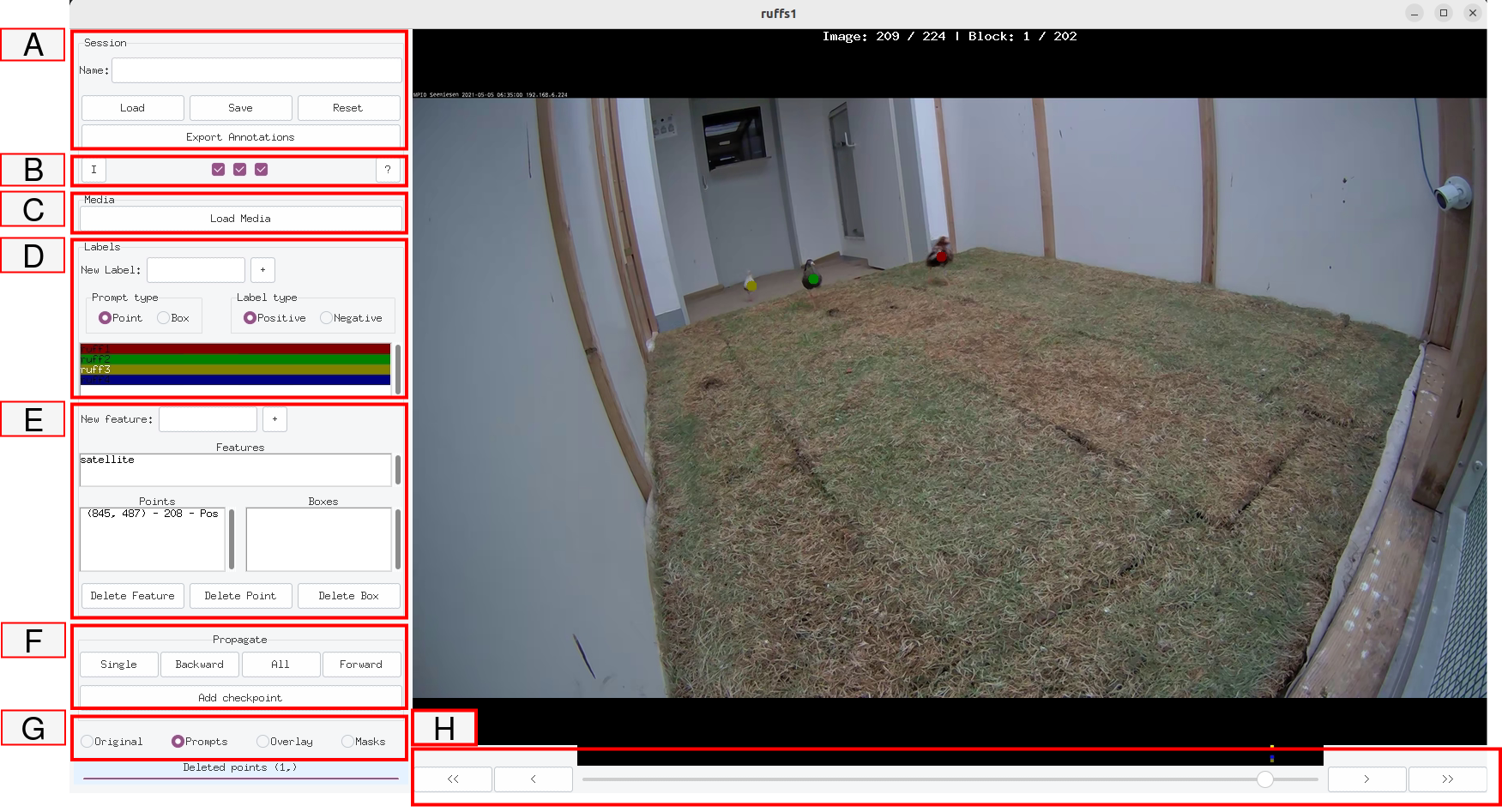}
    \caption{The graphical user interface is built up from the control panel (left), the canvas (right), and the slider (below the canvas). The control panel has the following modules: (A) loading and saving sessions (B) settings and information buttons (C) loading input media (D) label management (E) feature management (F) annotation propagation (G) toggle for visualization (H) frame slider.}
    \label{fig:GUI1}
\end{figure}

Upon launching the application, the user is presented with the graphical user interface (GUI), which is built up from three primary sections, the control panel, the canvas and the slider.
The layout of the GUI can be seen in Fig.~\ref{fig:GUI1}.

The control panel is located on the left-hand side, while the canvas displaying the visual content is on the right. 
A slider below the canvas enables easy mouse-based navigation through the currently loaded frames.

The GUI enables users to load data, partition it into subsequent frame sequences (blocks) if desired, annotate the content, and save the results.
To facilitate an efficient annotation workflow, the control panel is organized into the following interactive sections:

\begin{enumerate}
    \item[(A)] Session management allows users manage session configurations. Key functionalities include initializing new projects, resuming work via session loading, and accessing the central export utility to generate logs and final annotation outputs.

    \item[(B)] 
    Settings and information panel consolidates system monitoring, user support, and performance toggles.
    Resource awareness is provided via a system monitor that tracks CPU, RAM, and VRAM usage, providing real-time feedback and warnings about potential resource exhaustion. This is complemented by an information icon (''\texttt{I}``) for detailed hardware stats and a help icon (''\texttt{?}``) that provides access to the user guide and keyboard shortcuts (see Fig.~\ref{fig:popups}, Fig~\ref{fig:userguide1} and Fig~\ref{fig:userguide2} in \ref{sec:AppendixPopups}).
    Additionally, users can toggle three features: (i) \emph{Image caching}, which stores pre-rendered overlays to enhance interface responsiveness; (ii) \emph{Auto-prompting}, which facilitates block-to-block transitions by automatically generating point prompts; and (iii) \emph{Label display}, which projects semantic class names onto the masks' centers of mass for easier visual identification.

    \item[(C)] Loading input allows to import different media types for annotation, using the operating system’s default file browser. 
    The software supports both image directories and video files. 
    For image-based input, all images in the selected directory are loaded as a frame sequence. 
    At this stage, the user must also specify the block size (i.e., the number of frames per block), which determines the step granularity on the slider and timeline below the canvas. 
    The block size should be set based on available GPU memory—as higher frame counts increase memory demand—and the complexity of the visual content, including the number, similarity, and dynamics of classes and instances.

    \item[(D)] The label management section lets users create labels in the ``New label'' field and rename existing labels after double-clicking them. 
    Each label corresponds to a single object instance in the video. 
    Two prompt settings are provided: (i) the prompt type (point or box), and in case of point prompts (ii) the prompt \emph{sign} (positive or negative).
    Box prompts are constructed by two consecutive points marking the diagonally opposite vertices of a bounding region.
    Prompts are then applied directly on the canvas.
    Both point and box prompts can be deleted by selecting them in the control panel window and then pressing the corresponding delete button at the bottom of the Feature Management section.

    \item[(E)] Feature management allows users to assign descriptive, per-instance features throughout the sequence. 
    Feature values (e.g. position or state of an object) can be edited, and updates propagate forward.
    Feature values can also be deleted with the corresponding button, at the bottom of this section.

    \item[(F)] Annotation propagation
    provides the interface for mask generation and temporal tracking.
    Users can trigger \emph{single-frame generation} to compute masks exclusively for the current frame based on active prompts. 
    For temporal consistency, the interface offers \emph{directional propagation} controls: \emph{forward}, \emph{backward}, or \emph{bidirectional} (all). 
    The latter executes two separate unidirectional passes and concatenates the results. Additionally, users can set checkpoints to block propagation at specific frames, ensuring stable reference points.

    \item[(G)] In the visualization management section, users can choose among the available canvas visualization modes:
    (i) option ``original'' shows the input image; (ii) option ``prompts' overlays the prompts defined for the current frame (iii) option ``overlay'' visualizes the generated masks on the alpha channel of the image at $50\%$ opacity and (iv) option ``masks'' displays only the masks without the image content. 
    The latter two options are available only after masks have been generated for the current frame. The indices of the current frame and the active block are shown above the canvas.
\end{enumerate}

The canvas displays the currently processed frame. 
Users interact with the canvas via mouse clicks. 
When a label is selected and a frame is loaded, a click initiates a prompt-creation action; for box prompts, two clicks are required to define both opposing corners. 
Newly created prompts are visualized immediately on the canvas. 
Holding the \texttt{Ctrl} key while clicking an existing prompt selects both the prompt and its associated label (highlighted in yellow). 
This provides an alternative method for deletion and assists users in identifying the corresponding label within complex scenes.

The frame slider (denoted with (H) in Fig.~\ref{fig:GUI1}) facilitates both navigation and visual tracking of the annotation progress across the current part of the video (block).
Within the active block, the timeline displays several status indicators: the current cursor position is highlighted in yellow, while user-prompted frames are marked on a dark-blue track. 
A secondary, label-specific track indicates prompts for the currently selected object, rendered in its assigned color.
Propagation status is represented similarly: a green track denotes frames with generated masks, while a parallel track isolates outputs for the active label. 
Finally, checkpoint frames are marked in red. Collectively, these indicators allow for a rapid, multi-layer assessment of the annotation state.
Navigation is supported with step control buttons: \texttt{>} and \texttt{<} for individual frames, and \texttt{>>} and \texttt{<<} for switching between blocks."

\subsubsection*{Backend}
The primary objective of the annotation process is to generate instance-specific segmentation masks across all frames, guided by user-defined prompts. 

The backend consist of four main components (1) memory management of the input data (2) prompt and feature management (3) propagation and (4) saving results and logs.

\begin{figure}[ht]
    \centering
    \includegraphics[width=\linewidth]{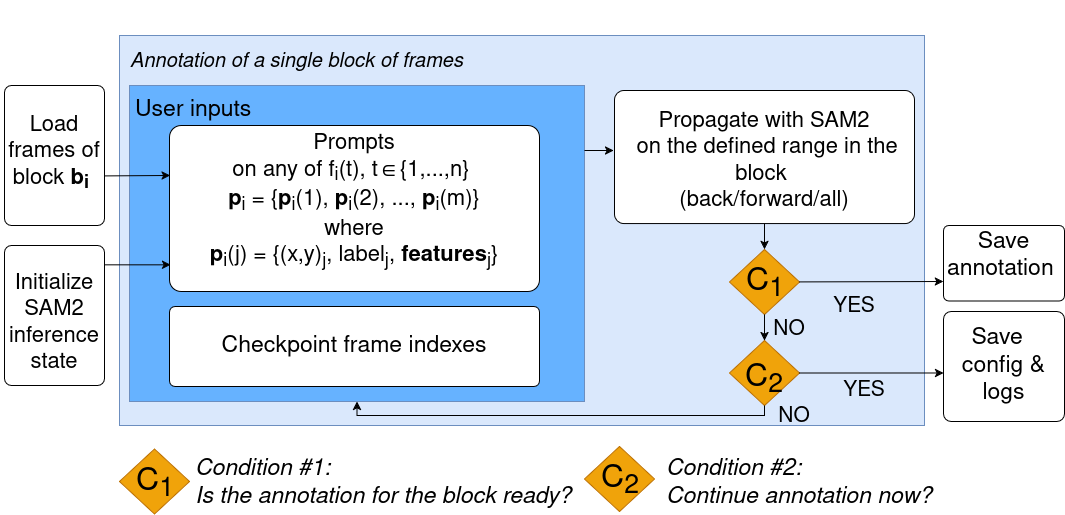}
    \caption{Control flow for annotating a single block: the input video is processed in blocks to enable efficient memory usage. Within each block, the user provides prompts on selected frames. We apply SAM2 to extend prompts within frames to masks, and propagate the masks to the remaining frames in the block. Finally, the system allows for saving the annotation configuration, log and the annotations.}
    \label{fig:control}
\end{figure}

Memory management enables efficient processing by keeping only the currently processed frames in memory. 
Therefore, the input video is partitioned into user-defined blocks of length $N$.
Each block is processed independently: edits within a block affect only frames in that block, except for labels and label features, which are shared globally across the entire video. 
Since SAM2 is initialized from an image directory rather than a video stream, switching blocks requires extracting the corresponding video segment and writing its frames to a temporary directory that is cleared at startup or upon session reset. 
Active frames are then copied to a dedicated temporary directory used to initialize the SAM2 inference state.\newline
The control flow annotating frames within a single block of frames is visualized in Fig.~\ref{fig:control}.\newline

Each target instance is associated with a \textit{label}, which is defined by a unique identification number (ID)
and a unique color value. 
Each label has a semantic name. 

To ensure architectural consistency within the configuration, the ID corresponds to the label's ordinal position in the registration sequence. 
Consequently, label deletion is prohibited to maintain the integrity of the indexing system. 
However, label names remain mutable to allow for typographical corrections or to reflect semantic shifts across different temporal blocks, provided that name uniqueness is preserved. 
To facilitate consistent visual tracking across the canvas and timeline, colors are mapped from the standard PASCAL VOC~\cite{Everingham2010VOC} colormap.
This choice, however, limits the number of labels to 256, the size of the colormap.
To ensure unambiguous instance identification during data export as image, the system maintains a strict identity mapping between the label ID and the colormap index; specifically, a label with ID $i$ is uniquely represented by the $i$-th entry of the colormap. 
This deterministic correspondence allows for the seamless recovery of individual instances from the exported segmentation masks.
\newline

The system implements the Segment Anything Model 2 (SAM2) prompt specification, supporting both point and box-based inputs. 
A box prompt (constructed by the diagonally opposite vertices of a bounding region) defines the spatial bounds of the target instance. Due to the architectural constraints of the underlying SAM2 model, box prompts are restricted to positive-only signals. 
Point prompts are categorized as either \emph{positive}, signifying the inclusion of a region within the instance mask, or \emph{negative}, indicating exclusion. It is important to note that negative prompts serve as local constraints and do not define a global background model, 
instead, they only guide instance representations by specifying which parts of an image do not belong to a specific label, following the internal logic of SAM2.

While the system supports multiple box prompts per label to accommodate complex scenarios (such as occlusions or semantic-style annotations involving disjoint regions or multiple objects in the scene) the current SAM2 propagation engine utilizes only the primary (first) box prompt for inter-frame mask projection. 
To enable iterative refinement, all prompts remain deletable. 
Within a given frame block, the model's inference state is aggregate: all prompts contribute to the latent representation regardless of their specific temporal placement within that block.
\newline

Instances may also have features. 
Features represent a set of additional descriptive parameters $\mathcal{V}$ (e.g., color, state) associated with an instance label. 
The addition or removal of a feature is a global operation across all frames $\mathbb{F}$, not constrained by block boundaries. 
Feature change is governed by the \textit{most recent change} rule. 
To illustrate the temporal propagation of values, consider a scenario where a feature is assigned a color value $C_1$ 
at $frame_k$ and is subsequently updated to 
another color $C_2$ 
at $frame_{k+n}$, with $k, n \in \mathbb{N}$ and $n > 0$. 
Assuming 
the value of a feature $f$ at any frame $t$ is denoted as $\text{val}_f(t)$, and no intermediate modifications exist in the log for the interval $(k, k+n)$, the value function for the feature is defined as follows:

\begin{equation}
\text{val}_f(t) = 
\begin{cases} 
C_1 & \text{if } t \in \{k, k+1, \dots, k+n-1\} \\
C_2 & \text{if } t \geq k+n 
\end{cases}
\end{equation}

This implies that the state is persistent on the discrete interval $[k, k+n)$, and the value change at $t = k+n$ is inclusive, meaning the new value takes effect immediately at the frame of modification. 
\newline

The propagation module is responsible for the spatio-temporal extension of sparse user prompts into dense segmentation masks across the video timeline.
This process is orchestrated by a specialized wrapper class around the Segment Anything Model 2 (SAM2), which encapsulates directional inference, state management, and memory optimization.
To maintain interface responsiveness, all propagation tasks are executed on a dedicated worker thread, decoupling the computationally intensive inference from the graphical user interface.

Prompt selection for propagation follows the \textit{closest change} principle previously established for feature modifications. 
For a given label $l$ at frame $t$, the propagation engine identifies the most recent frame $s$ containing user-defined prompts. Formally, let $P_l(t)$ be the set of prompts (points or boxes) associated with label $l$ at frame $t$. During propagation, the active prompt set $P^*_l(t)$ is defined as:

\begin{equation}
P^*_l(t) = P_l(s), \quad \text{where } s = \max \{k \in \mathbb{N} \mid k \le t \wedge P_l(k) \neq \emptyset \}
\end{equation}

If no such $s$ exists within the current temporal scope, label $l$ is excluded from the propagation cycle. For single-frame mask generation, the engine strictly considers $s = t$. 
Consistent with the propagation requirements of SAM2, while multiple box prompts may be assigned to a label to accommodate disjoint regions, only the primary (first) box prompt is utilized for inter-frame mask propagation.\newline

\comment{The system supports three propagation modes: \textit{unidirectional} (forward or backward), \textit{bidirectional} (implemented as two separate unidirectional passes with concatenated results), and \textit{singular} (restricted to mask generation on the current frame). 
Regardless of the mode, propagation is subject to a set of temporal constraints $\mathcal{B} = \{t_{curr}, t_{edge}, t_{check}\}$, defined as follows:
\begin{itemize}
    \item $t_{curr}$: The initiation frame, constituting one or both limits depending on direction.
    \item $t_{edge}$: The boundaries of the current temporal block, serving as terminal limits.
    \item $t_{check}$: User-defined \textit{checkpoints} that function as immutable barriers. Propagations initiated on one side of a checkpoint cannot cross or overwrite predictions on the opposing side, enabling a ``lock-and-refine'' workflow.
\end{itemize}}
The system facilitates three distinct propagation modes: \textit{unidirectional} (forward or backward), \textit{bidirectional} (realized as two independent unidirectional passes with concatenated results), and \textit{singular} (restricted to mask generation on the current frame). Regardless of the selected mode, propagation is strictly governed by a set of temporal constraints $\mathcal{B} = \{t_{curr}, t_{edge}, t_{check}\}$. In this framework, $t_{curr}$ refers to the initiation frame that establishes the temporal origin; $t_{edge}$ represents the boundaries of the active temporal block, serving as the terminal limits for inference; and $t_{check}$ denotes user-defined \textit{checkpoints}. These checkpoints function as immutable barriers, ensuring that any propagation initiated on one side of the boundary can neither traverse nor overwrite predictions on the opposite side. This mechanism effectively enables a ``lock-and-refine'' workflow, allowing users to preserve validated segments while iteratively updating others.\newline

To mitigate the substantial memory overhead of propagation associated with SAM2's attention mechanisms in extended sequences, we implement a sliding-window strategy inspired by community-driven optimizations~\cite{aza1200_sam2_issue196_2024,aendrs_sam2_issue264_2024}. 
Specifically, the system explicitly discards stored latent frame data older than 16 frames relative to the current inference point.
Furthermore, in SAMannot we reinitialize the SAM2 inference state for each propagation cycle within a block. 
This ensures memory fragmentation is minimized while preserving the ability to iteratively correct earlier annotations. 
Asynchronous frame loading is employed during state initialization, with retention disabled for these pre-loaded frames to further reduce the RAM footprint.

To streamline the transition between blocks, we introduce an \textit{auto-prompting} mechanism. 
Upon completion of a block, the system automatically generates point prompts for the initiation of the subsequent block. 
For each label $l$, the generated mask $M_l$ is reduced to its topological skeleton $\mathcal{S}(M_l)$. 
Point prompts are then strategically placed at endpoints and junctions of $\mathcal{S}(M_l)$ using pixel-neighborhood connectivity analysis.

To account for rapid spatial shifts (common in e.g. animal behavior tracking) the system employs a one-frame temporal lookahead. The wrapper class processes frame $t_{end}+1$ of the succeeding block during propagation. 
The resulting mask is used solely for auto-prompt generation and is not exposed to the user, ensuring that the prompts for the new block accurately reflect the instance's position at the start of the next interval.
\newline

The fourth component of the backend architecture is dedicated to saving results and logs, providing a robust framework for both dataset generation and state persistence.
This module facilitates the generation of research-ready datasets while ensuring the long-term persistence of the application's internal state. 
All outputs are organized into session-specific subdirectories, exclusively containing frames with valid annotations --- whether manually produced or model-generated. Human input prompts are explicitly tagged, allowing for a clear distinction between manual annotations and inference-based data.

Exported data are categorized into pixel-level masks and structured tabular metadata.
Segmentation masks are provided in both lossless PNG and standardized YOLO formats. 
To facilitate immediate integration into downstream machine learning pipelines, these artifacts are co-located with their corresponding raw image frames. 
The PNG masks maintain a deterministic color-to-label mapping, ensuring that pixel values directly correlate with the established label IDs.

The system can generate tabular metadata and interaction logs documenting the interplay between user input and model inference. 
For each frame and label, the export includes the precise coordinates of all prompts $P_l$. 
Additionally, the system calculates and exports the centroid (center of mass) for each generated mask. 
These centroids serve as proxies for the instance's canonical position, representing the predicted target for subsequent user interactions. 
A comprehensive mapping schema between label identifiers, semantic names, and colormap indices is included to ensure alignment.

To support asynchronous workflows, the entire application state can be serialized using \emph{pickle} package in Python. 
The resulting session file encapsulates the complete operational context, including temporal annotation logs, feature modifications, visualization parameters, and the internal states of GUI components. 
This enables the exact restoration of the environment, allowing for iterative refinement of long-form video sequences over multiple sessions.

\section*{Quality control}

The software development process incorporated regular code reviews and continuous assessment of potential improvements. 

The reliability and cross-platform compatibility of SAMannot have been verified through functional testing on Ubuntu 22.04 LTS and Microsoft Windows 11. 
SAMannot behaved as expected on all platforms.

The software undergoes a rigorous validation process prior to each release, structured as follows:
\begin{itemize}
    \item \textbf{Functional and Usability Testing:} Usability assessments are conducted in collaboration with researchers at the Max Planck Institute for Biological Intelligence to ensure the workflow meets the requirements of real-world biological research.
    These tests comprise a suite of standard sessions using short sample datasets to emulate diverse annotation workflows, including both propagated and non-propagated tasks. Collectively, they cover all core software functionalities, media handling, user interaction (global and per-frame), propagation, rendering, and data export.
    
    \item \textbf{Installation Verification:} To enable rapid verification of a successful deployment, the repository includes a sample workflow (comprising a short video sequence and two pre-configured session files), allowing the users to execute a baseline propagation and export cycle, comparing the output against reference masks to confirm correct environment setup and GPU acceleration.
    
    \item \textbf{Performance Benchmarking:} The accuracy of the integrated SAM2-based propagation has been validated using subsets of the DAVIS~\cite{Perazzi2016} benchmark.
    These tests confirm that the optimized wrapper maintains high precision while maintaining a low memory footprint compared to the baseline SAM2 implementation.
    \item \textbf{Maintenance and Support:} 
    During development, biologists and other collaborators using the software in their projects provided the technical issue reports and feature requests. Following release, this process will transition to the GitHub issue tracker to enable transparent, community-facing support and maintenance.
\end{itemize}

\textbf{Performance evaluation}

Measurements were performed on a desktop computer with Ubuntu 22.04, equipped with an AMD Ryzen Threadripper Pro 5955WX CPU, NVIDIA RTX 4090 GPU with 24GB of VRAM, and 512GB of RAM.

To quantitatively evaluate the segmentation accuracy of SAMannot, we conducted an evaluation using the semi-supervised \emph{TrainVal} subset of the DAVIS 2017 benchmark~\cite{DAVIS} and the validation set of the LVOS dataset~\cite{LVOS}. 
We randomly selected folders from both datasets to ensure a diverse benchmark covering various image characteristics and instance counts, and different video lengths.
In total, we evaluated 22 sequences from the DAVIS dataset, comprising $1\,681$ frames with an average length of 76 frames per video. Additionally, 5 sequences were selected from the LVOS dataset, totaling $3\,783$ frames with an average video length of $757$ frames.

\begin{figure}[ht]
    \centering
    \includegraphics[width=1.0\linewidth]{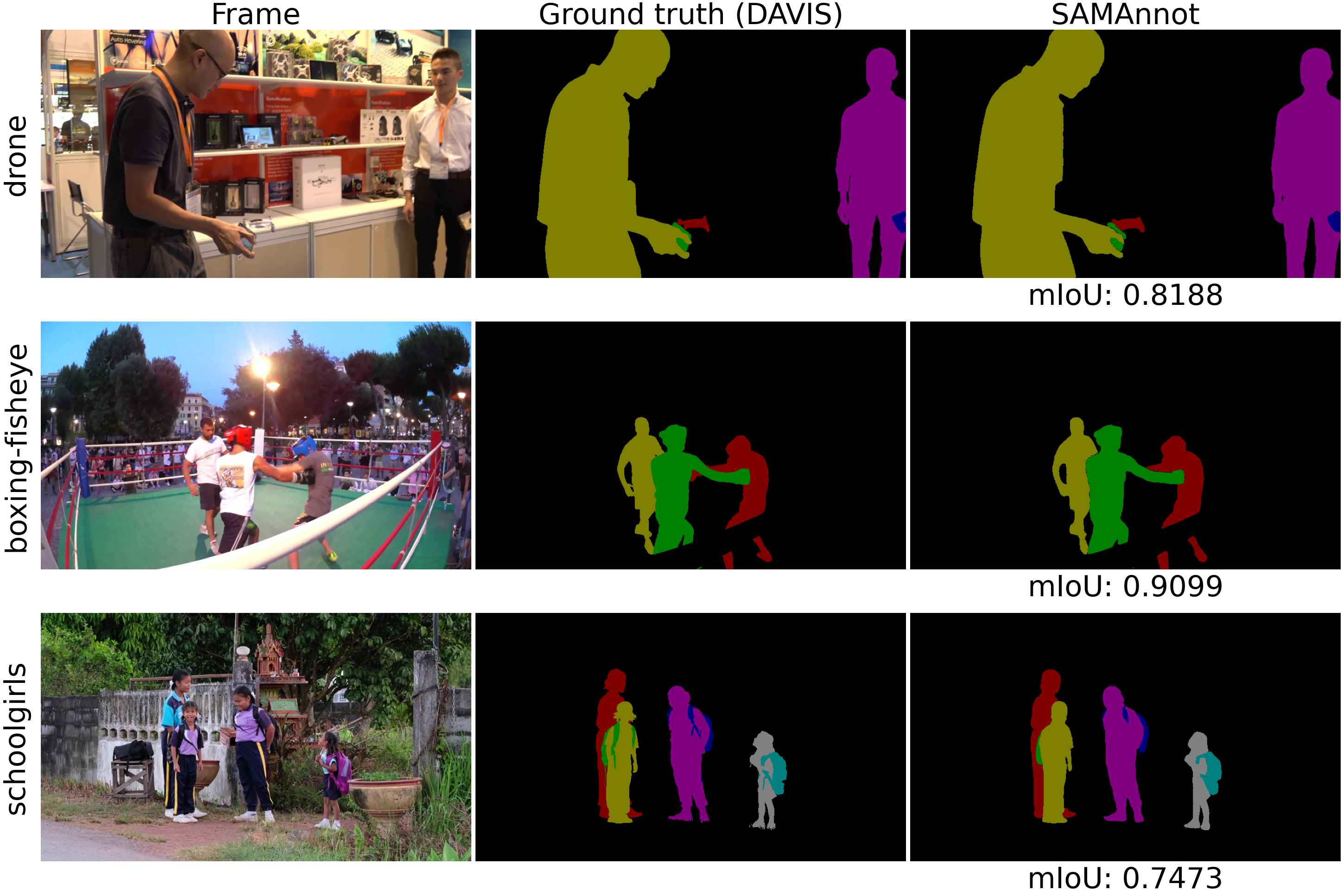}
    \caption{Qualitative examples of segmentation results on images from the DAVIS 2017 dataset. The columns display the original video frame (left), the ground truth (middle), and the masks predicted by SAMannot (right).}
    \label{fig:grid}
\end{figure}

Each sequence was evaluated using three metrics: Intersection over Union (IoU)~\cite{iou} that rigorously penalizes boundary misalignments and missed fine details, 
the Dice coefficient (or F1-score)~\cite{f1score} measuring how much the predicted and ground-truth masks overlap, normalized by their sizes, 
and Pixel Accuracy~\cite{randindex} is included to provide a baseline for total classification correctness, although its interpretation remains secondary due to its inherent sensitivity to class imbalance in high-resolution video frames.
Table~\ref{tab:DAVISres} and Table~\ref{tab:LVOSres} summarize the quantitative results, demonstrating that SAMannot achieves high-quality annotations comparable to the ground truth.
For a few cases with lower IoU, we show examples in Fig.~\ref{fig:grid} to demonstrate that the segmentations remain qualitatively accurate despite the lower scores.
Additional visualizations are provided in ~\ref{app:visAnnot}, Fig.~\ref{fig:totalgrid}.

The results indicate that segmentation accuracy is primarily driven by the visual characteristics of the targets rather than the absolute number of instances.

\begin{figure}[ht!]
    \centering
    \begin{minipage}[t]{0.32\textwidth}
        \centering
        \includegraphics[width=\linewidth]{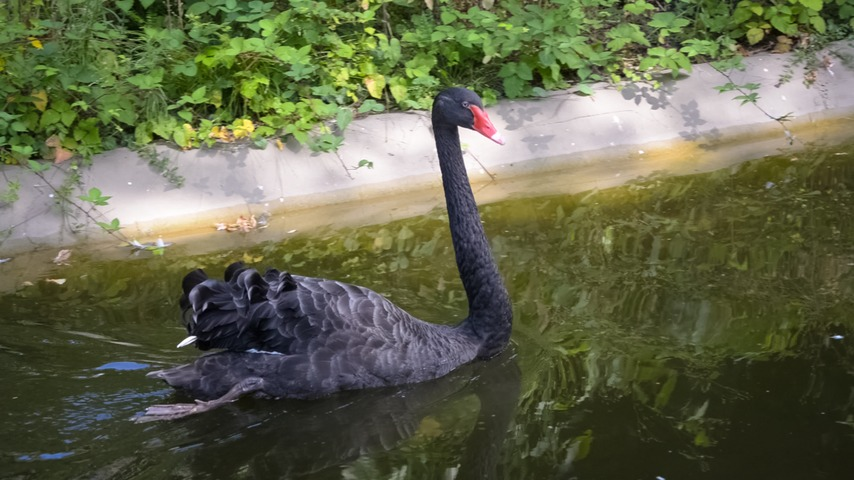}
        \small Original frame
    \end{minipage}\hfill
    \begin{minipage}[t]{0.32\textwidth}
        \centering
        \includegraphics[width=\linewidth]{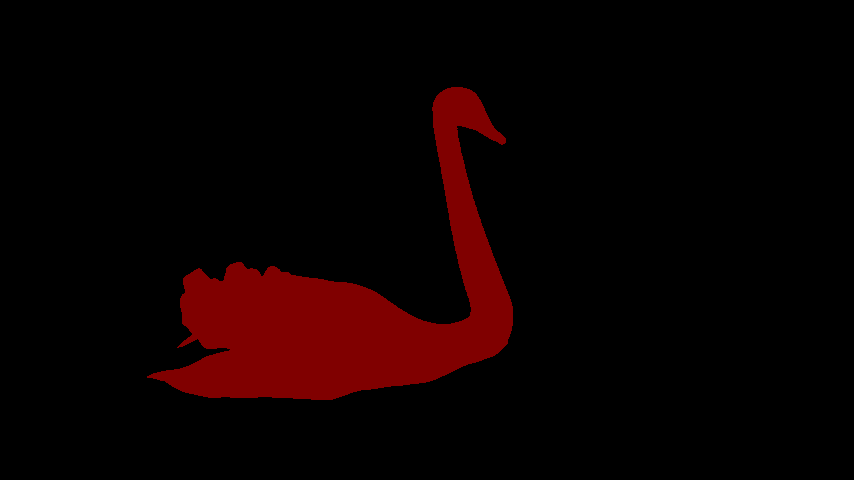}
        \small DAVIS ground truth
    \end{minipage}\hfill
    \begin{minipage}[t]{0.32\textwidth}
        \centering
        \includegraphics[width=\linewidth]{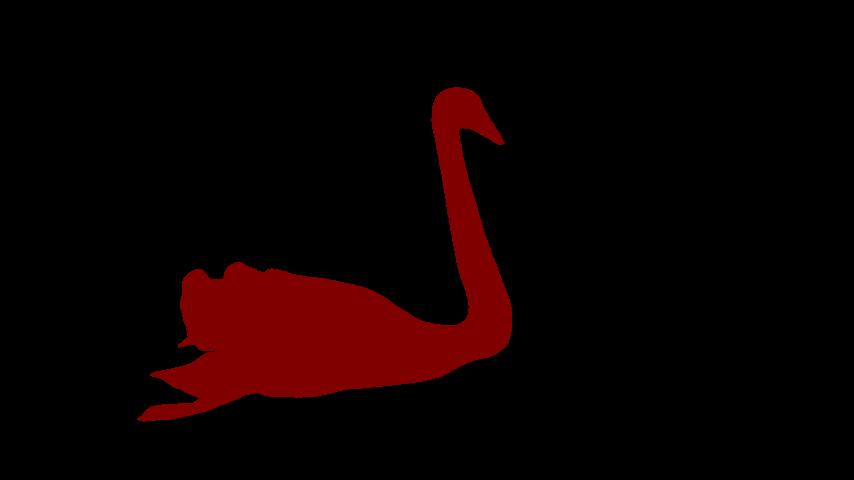}
        \small SAMannot prediction
    \end{minipage}

    \caption{Comparison of semantic segmentation boundaries. From left to right: original frame of the \emph{Blackswan} sequence, official DAVIS 2017 ground truth, and SAMannot prediction. Note the discrepancy regarding the swan's feet: while the ground truth excludes them, SAMannot correctly identifies these regions as part of the semantic instance. Such differences contribute to a lower measured Mean IoU and Mean Dice, despite the model providing a more anatomically complete segmentation.}
    \label{fig:kacsa}
\end{figure}

Qualitative analysis revealed certain discrepancies in edge cases. 
As illustrated in Fig.~\ref{fig:kacsa}, these differences often arise from semantic ambiguity: 
in some instances, the DAVIS ground truth excludes anatomically or functionally connected parts of an object, or merges distinct entities that SAMannot correctly identifies as separate semantic instances.

\begin{table}[h!]
\centering
\resizebox{\linewidth}{!}{%
\begin{tabular}{l c c c>{\columncolor{gray!15}}c !{\color{white}\vrule} >{\columncolor{gray!15}}c c} 
\toprule
\makecell{\textbf{Sequence}\\\textbf{name}} & \makecell{\textbf{Frames}\\\textbf{(\#)}} & \makecell{\textbf{Instances}\\\textbf{(\#)}} & \makecell{\textbf{All masks}\\\textbf{(\#)}} & \makecell{\textbf{Mean}\\\textbf{IoU}}  & \makecell{\textbf{Mean}\\\textbf{Dice}} & \makecell{\textbf{Pixel}\\\textbf{Acc.}} \\ \midrule
Rhino            & 90 & 1 & 90 & 0.9807  & 0.9902 & 0.9968 \\
Cows             & 104& 1 & 104 & 0.9710  & 0.9853 & 0.9966 \\
Bear             & 82 & 1& 82 & 0.9703  & 0.9849 & 0.9966 \\
Camel            & 90 & 1& 90  & 0.9691  & 0.9843 & 0.9962 \\
Dog              & 60 & 1& 60 & 0.9640  & 0.9817 & 0.9960 \\
Breakdance       & 84 & 1& 84 & 0.9593  & 0.9792 & 0.9963 \\
{Breakdance-flare} & 71 & 1& 71 & 0.9586  & 0.9789 & 0.9974 \\
Tuk-tuk          & 59 & 3& 177 & 0.9515  & 0.9747 & 0.9783 \\
Blackswan        & 50 & 1& 50 & 0.9506  & 0.9746 & 0.9950 \\
Cat-girl         & 89 & 2& 178 & 0.9460  & 0.9722 & 0.9838 \\
Night-race       & 46 & 2& 83 & 0.9445  & 0.9656 & 0.9973 \\
Train            & 80 & 4& 320 & 0.9290  & 0.9631 & 0.9877 \\
Bus              & 80 & 1& 80 & 0.9295  & 0.9626 & 0.9883 \\
Classic-car      & 63 & 3& 189 & 0.9265  & 0.9579 & 0.9879 \\
Color-run        & 84 & 3& 217 & 0.9252  & 0.9579 & 0.9695 \\
Boxing-fisheye   & 87 & 3& 261 & 0.9099  & 0.9522 & 0.9948 \\
Bike-packing     & 69 & 2& 138 & 0.9026  & 0.9482 & 0.9825 \\
Pigs             & 79 & 3& 237 & 0.8914  & 0.9278 & 0.9885 \\
Boat             & 75 & 1& 75 & 0.8243  & 0.9036 & 0.9882 \\
Sheep            & 68 & 5& 340 & 0.8361  & 0.8945 & 0.9951 \\
Drone            & 91 & 4& 298 & 0.8188  & 0.8649 & 0.9627 \\
Schoolgirls      & 80 & 7& 560 & 0.7473  & 0.8214 & 0.9896 \\
\midrule
\textbf{Average} & \textbf{76} & -- & \textbf{172} 
& \textbf{0.9185} & \textbf{0.9512} & \textbf{0.9893} \\ 
\textbf{Std dev.} &	\textbf{14.37}& -- &	\textbf{125.27}	& \textbf{0.0606} &	\textbf{0.0436}	& \textbf{0.0093} \\
\bottomrule
\end{tabular}
}
\caption{Quantitative evaluation of SAMannot on a subset of the DAVIS 2017 train-val dataset (480p resolution). The metrics represent the mean Intersection over Union (IoU), Dice coefficient, and Pixel Accuracy for each sequence. 
}
\label{tab:DAVISres}
\end{table}

The LVOS dataset presents a high level of difficulty due to demanding conditions, such as frequent occlusions, substantial motion blur, and long-term object reappearances, representing the ''hard`` end of the complexity spectrum.\newline

\begin{table}[h!]
\centering
\resizebox{\linewidth}{!}{%
\begin{tabular}{l c c c >{\columncolor{gray!15}}c !{\color{white}\vrule} >{\columncolor{gray!15}}c c} 
\toprule
\makecell{\textbf{Sequence}\\\textbf{name}} & \makecell{\textbf{Frames}\\\textbf{(\#)}} & \makecell{\textbf{Instances}\\\textbf{(\#)}} & \makecell{\textbf{All masks}\\\textbf{(\#)}} & \makecell{\textbf{Mean}\\\textbf{IoU}}  & \makecell{\textbf{Mean}\\\textbf{Dice}} & \makecell{\textbf{Pixel}\\\textbf{Acc.}} \\ \midrule
3bvEjhOT  & 461  & 2 & 896 & 0.9585  & 0.9778 & 0.9967 \\
7K7WVzGG  & 617  & 2 & 1262 & 0.8030  & 0.8391 & 0.9989 \\
cUD1dwuP   & 793 & 4 & 2723 & 0.9282  & 0.9588 & 0.9969 \\
EWCZAcdt  & 1412 & 2 & 2056 & 0.8333  & 0.9010 & 0.9993 \\ 
HYSm91eM  & 500  & 10 & 4992 & 0.7997  & 0.8452 & 0.9934 \\
\midrule
\textbf{Average} & \textbf{757} & -- & \textbf{2386}
& \textbf{0.8645} & \textbf{0.9044} & \textbf{0.9970} \\ 
\textbf{Std dev.} &	\textbf{388.45}	& --	& \textbf{1619.97}	&	\textbf{0.0739}	&	\textbf{0.0635}	&	\textbf{0.0023}	\\
\bottomrule
\end{tabular}
}
\caption{Quantitative evaluation of SAMannot on a subset of the LVOS dataset~\cite{LVOS}. The metrics represent the mean Intersection over Union (IoU), Dice coefficient, and Pixel Accuracy for each sequence.}
\label{tab:LVOSres}
\end{table}

\begin{figure}[H]
    \centering
    \begin{subfigure}[b]{0.19\linewidth}
        \includegraphics[width=2.85cm, height=1.8cm]{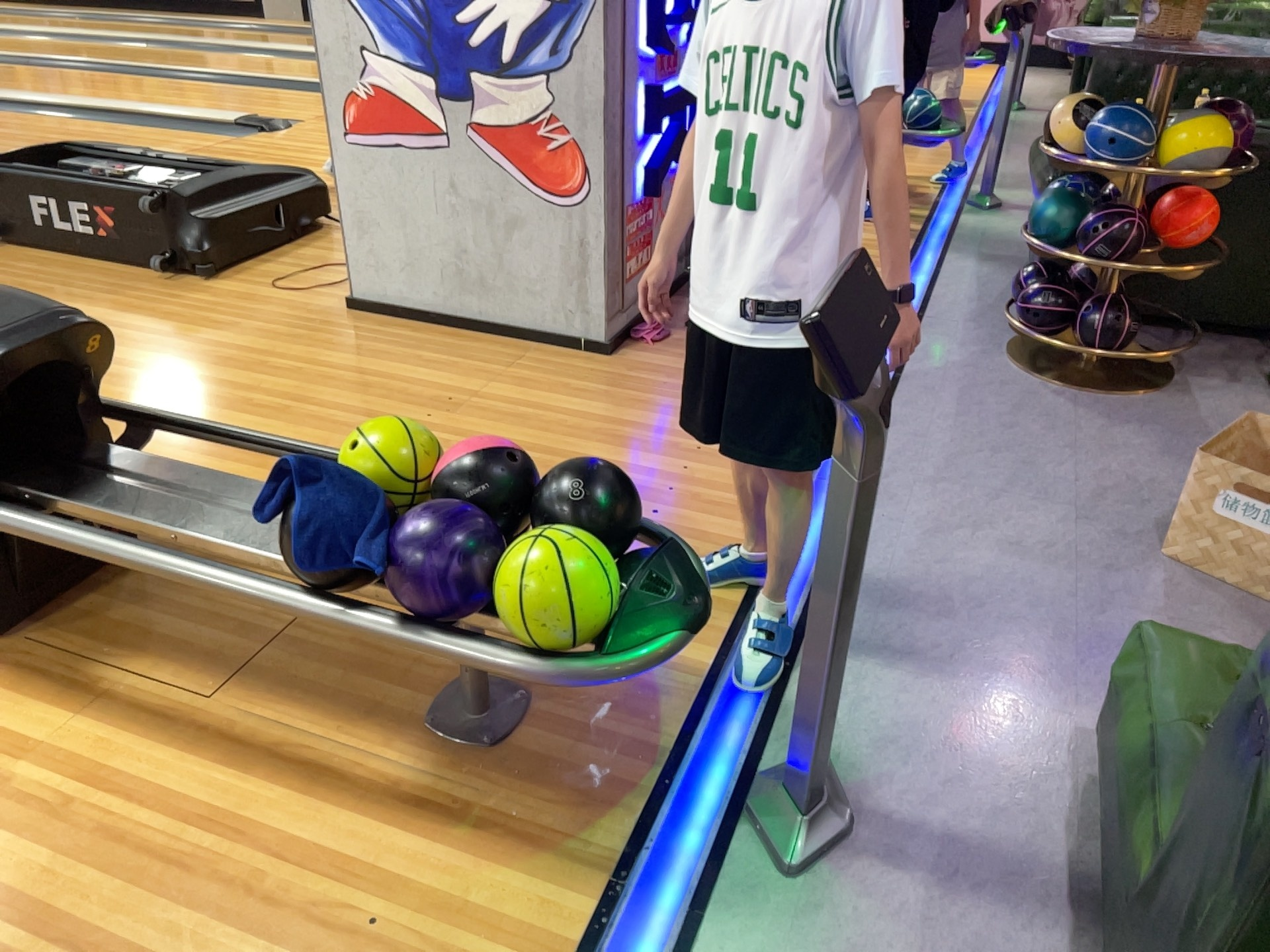}
        \caption*{3bvEjhOT}
        \label{subfig:lvos1}
    \end{subfigure}
    \hfill
    \begin{subfigure}[b]{0.19\linewidth}
        \includegraphics[width=\linewidth,height=1.8cm]{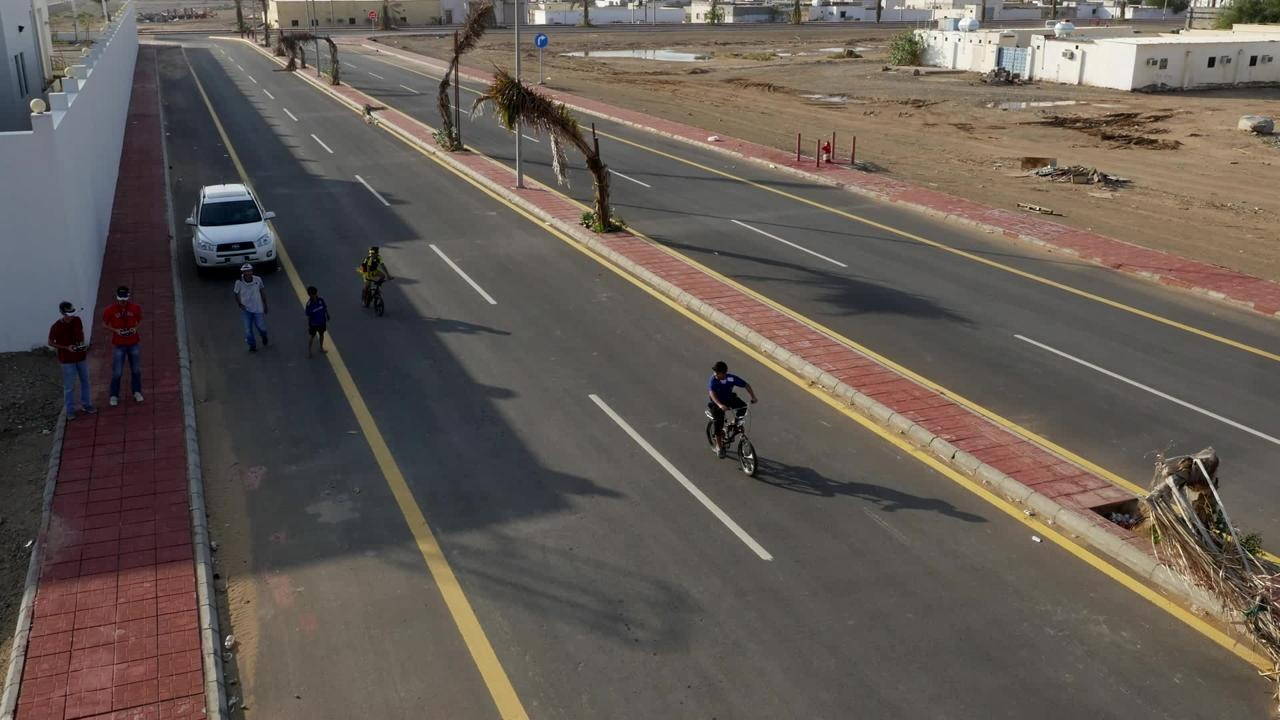}
        \caption*{7K7WVzGG}
        \label{subfig:lvos2}
    \end{subfigure}
    \hfill
    \begin{subfigure}[b]{0.19\linewidth}
        \includegraphics[width=\linewidth,height=1.8cm]{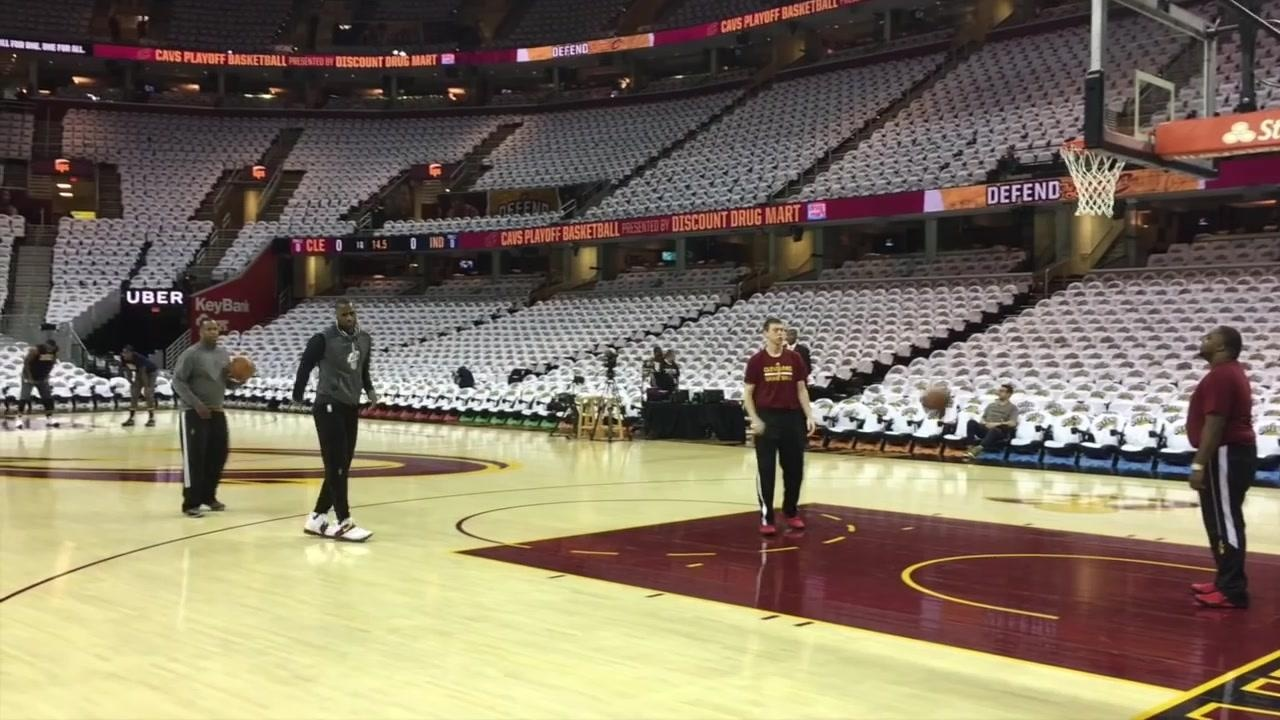}
        \caption*{cUD1dwuP}
        \label{subfig:lvos3}
    \end{subfigure}
    \hfill
    \begin{subfigure}[b]{0.19\linewidth}
        \includegraphics[width=\linewidth,height=1.8cm]{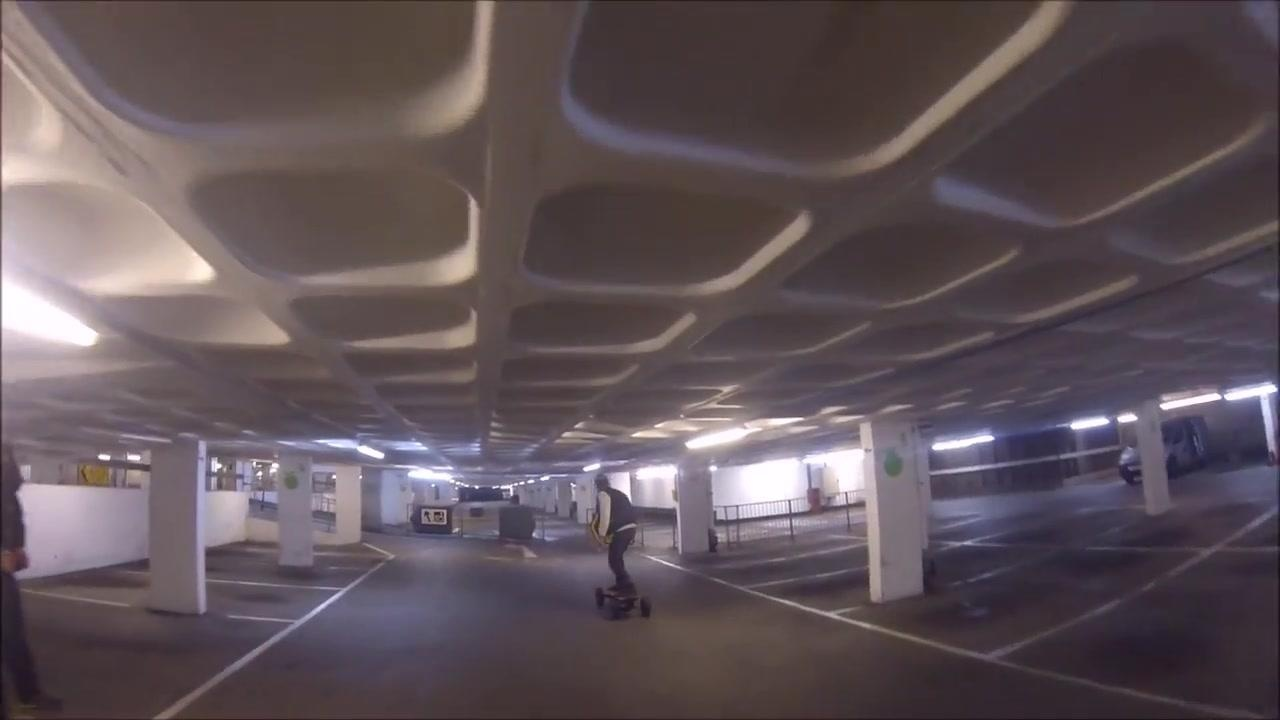}
        \caption*{EWCZAcdt}
        \label{subfig:lvos4}
    \end{subfigure}
    \hfill
    \begin{subfigure}[b]{0.19\linewidth}
        \includegraphics[width=\linewidth,height=1.8cm]{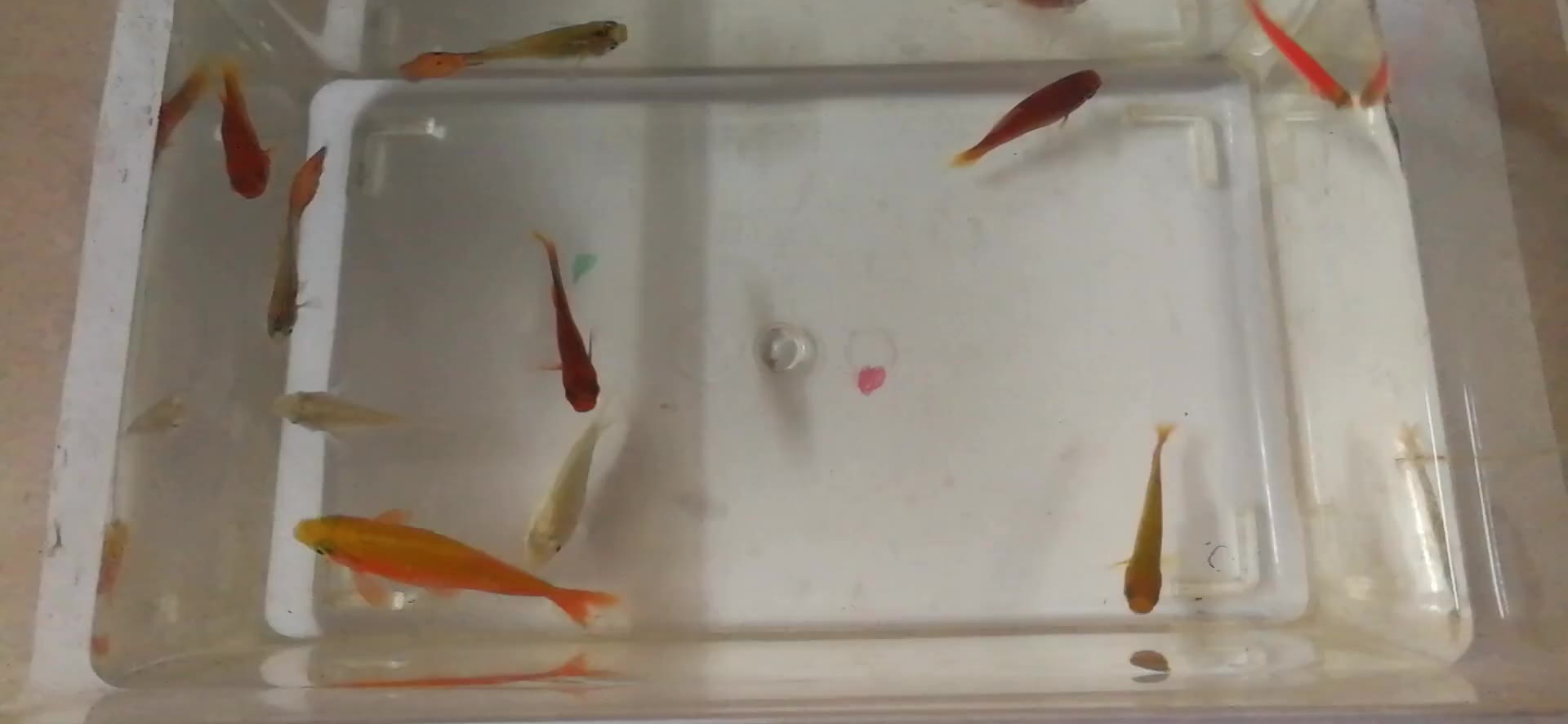}
        \caption*{HYSm91eM}
        \label{subfig:lvos5}
    \end{subfigure}

    \caption{Illustrative frames from the analyzed LVOS sequences, demonstrating the visual diversity of the dataset, for the performance metrics.}
    \label{fig:LVOS_examples}
\end{figure}

To validate our memory optimization strategies, we monitored VRAM utilization across multiple annotation sessions (see Fig.~\ref{fig:DAVIS_examples} and Fig.~\ref{fig:LVOS_examples} for sample frames and Table~\ref{tab:performanceDAVIS} and Table~\ref{tab:performanceLVOS} for evaluation metrics).
GPU utilization was monitored using the \texttt{nvidia-smi} utility (NVIDIA System Management Interface)~\cite{nvidiasmi}. Occupancy data were recorded to a CSV file at a sampling frequency of $5Hz$ (one sample every 0.2 seconds) to track resource consumption alongside session timestamps.
The results demonstrate that by partitioning the inference process into manageable blocks, the software maintains a stable memory footprint, preventing the typical VRAM overflow associated with long-sequence processing in baseline foundation models.

Annotation times were recorded for each sequence to give an insight on operational efficiency.
Actual performance is highly sensitive to scene complexity and the experience of the annotator thus, these durations reflect an 
indicative approximation 
rather than a benchmark.

\newcommand{\mysize}{0.19}
\begin{figure}[H]
    \centering
    \begin{subfigure}[b]{\mysize\linewidth}
        \includegraphics[width=\linewidth]{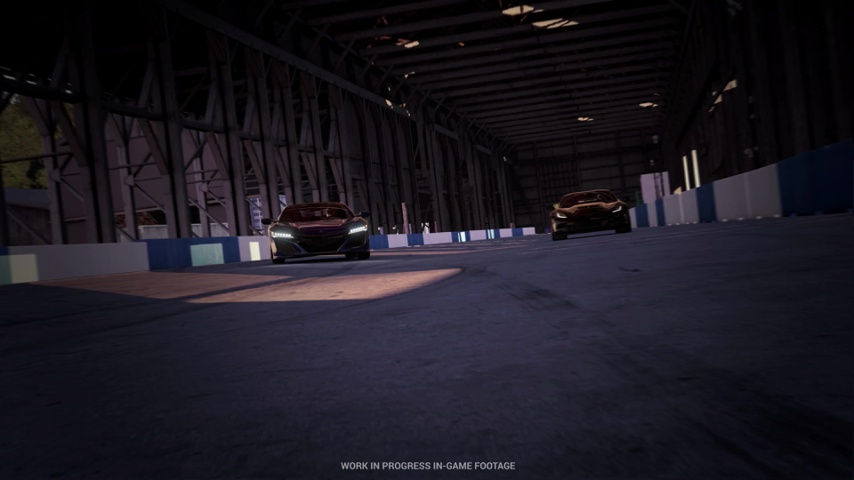}
        \caption*{night-race}
        \label{subfig:nightrace}
    \end{subfigure}
    \hfill
    \begin{subfigure}[b]{\mysize\linewidth}
        \includegraphics[width=\linewidth]{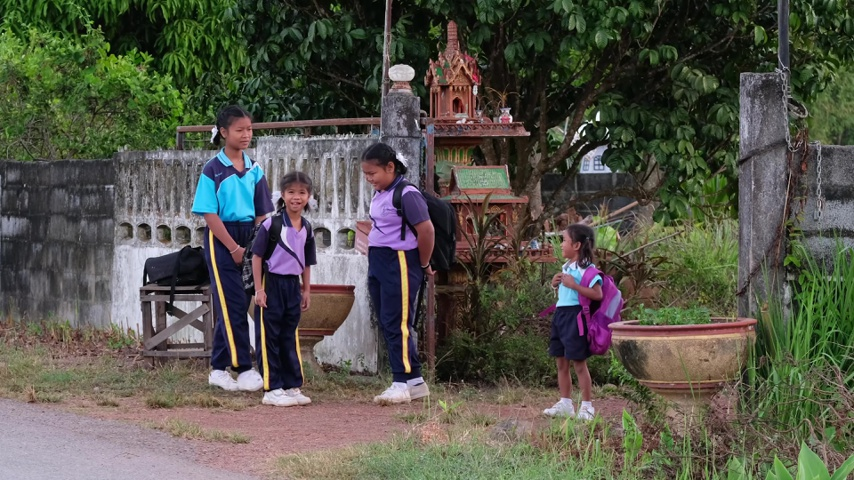}
        \caption*{schoolgirls}
        \label{subfig:schoolgirls}
    \end{subfigure}
    \hfill
    \begin{subfigure}[b]{\mysize\linewidth}
        \includegraphics[width=\linewidth]{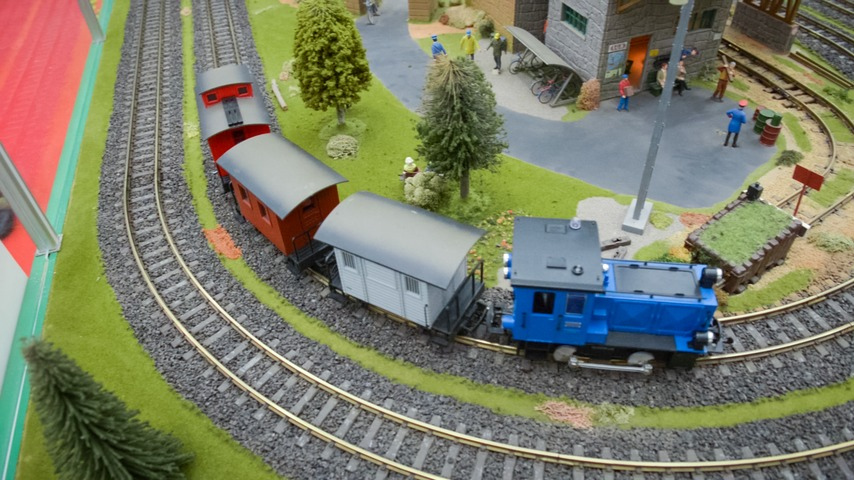}
        \caption*{train}
        \label{subfig:train}
    \end{subfigure}
    \hfill
    \begin{subfigure}[b]{\mysize\linewidth}
        \includegraphics[width=\linewidth]{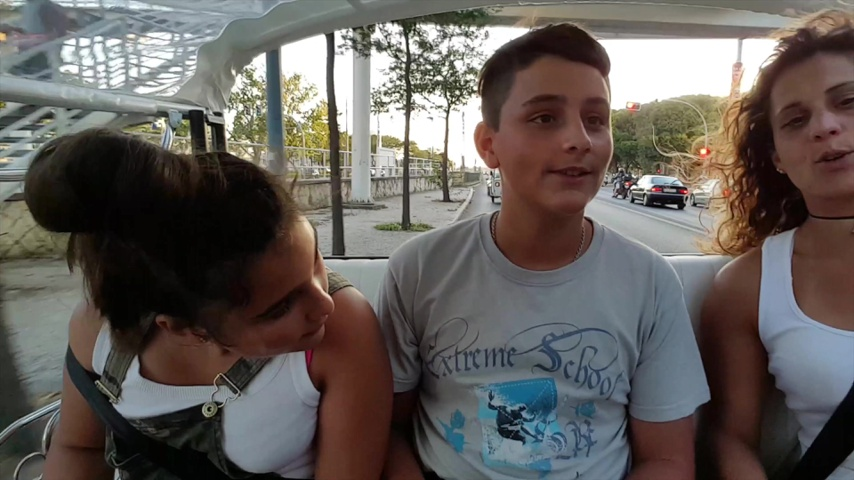}
        \caption*{tuk-tuk}
        \label{subfig:tuktuk}
    \end{subfigure}
    \hfill
    \begin{subfigure}[b]{\mysize\linewidth}
        \includegraphics[width=\linewidth]{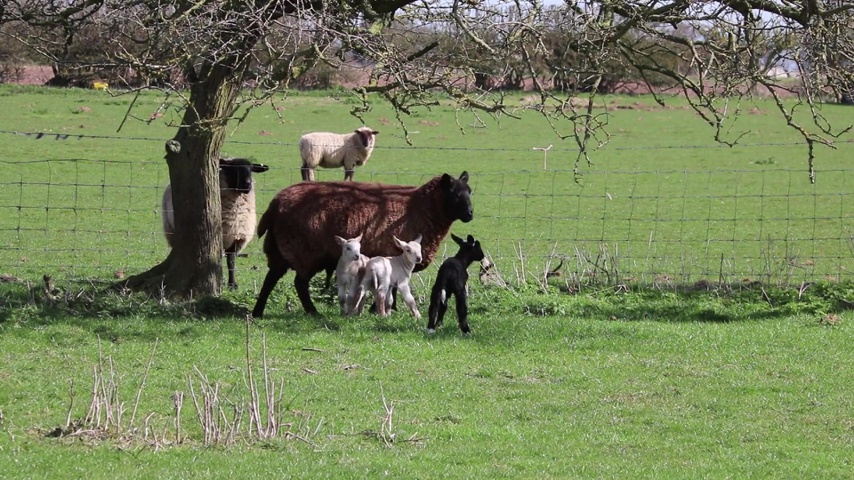}
        \caption*{sheep}
        \label{subfig:sheep}
    \end{subfigure}
    \caption{Illustrative frames from the analyzed DAVIS sequences for the performance metrics.}
    \label{fig:DAVIS_examples} 
\end{figure}

\begin{table}[h]
\centering
\small
\setlength{\tabcolsep}{4pt} 
\resizebox{\linewidth}{!}{%
\begin{tabular}{lcccc>{\columncolor{gray!15}}cc}
\hline
\textbf{Video} & \textbf{Inst.} & \textbf{Frames} & \textbf{Duration} & \textbf{VRAM$_{min}$} & \textbf{VRAM$_{max}$} & \textbf{$\Delta$VRAM} \\
\textbf{Name} & \textbf{(\#)} & \textbf{(\#)} & \textbf{(mm:ss)} & \textbf{(MiB)} & \textbf{(MiB)} & \textbf{(MiB)} \\ \hline
night-race & 2 & 46 & 0:47& 1503 & 2354 & 851 \\
schoolgirls & 7 & 80 & 5:09 & 1536 & 2898 & 1362 \\
train & 4 & 80 & 4:09 & 1512 & 2711 & 1199 \\
tuk-tuk & 3 & 59 & 1:33 & 1509 & 2669 & 1160 \\
sheep & 5 & 68 & 1:35 & 1519 & 2711 & 1192 \\
\end{tabular}
}
\caption{Performance metrics and resource utilization during video annotation on videos from the DAVIS 2017 dataset~\cite{DAVIS}. \textit{Duration} 
encompasses
label definition, annotation, and the final data export.}
\label{tab:performanceDAVIS}
\end{table}

\begin{table}[H]
\centering
\small
\setlength{\tabcolsep}{4pt} 
\resizebox{\linewidth}{!}{%
\begin{tabular}{lcccc >{\columncolor{gray!15}}cc}
\hline
\textbf{Video} & \textbf{Inst.} & \textbf{Frames} & \textbf{Duration} & \textbf{VRAM$_{min}$} & \textbf{VRAM$_{max}$} & \textbf{$\Delta$VRAM} \\
\textbf{Name} & \textbf{(\#)} & \textbf{(\#)} & \textbf{(mm:ss)} & \textbf{(MiB)} & \textbf{(MiB)} & \textbf{(MiB)} \\ \hline
3bvEjhOT & 2 & 461 & 7:42 & 1521 & 2713 & 1192 \\
7K7WVzGG & 2 & 617 & 8:30 & 1523 & 2687 & 1164 \\
cUD1dwuP & 4 & 793 & 15:49 & 2112 & 2763 & 651 \\
EWCZAcdt & 2 & 1412 & 19:25 & 2150 & 2843 & 693 \\
HYSm91eM & 10 & 500 & 16:56 & 1530 & 2868 & 1338 \\
\end{tabular}
}
\caption{Performance metrics and resource utilization during video annotation on videos from the LVOS~\cite{LVOS} dataset. \textit{Duration} 
encompasses
label definition, annotation, and the final data export.}
\label{tab:performanceLVOS}
\end{table}

\textbf{Ground-truth inconsistencies}\newline

\begin{figure}[H]
    \centering
    \begin{subfigure}[b]{0.3\linewidth} 
        \centering
        \includegraphics[height=6cm]{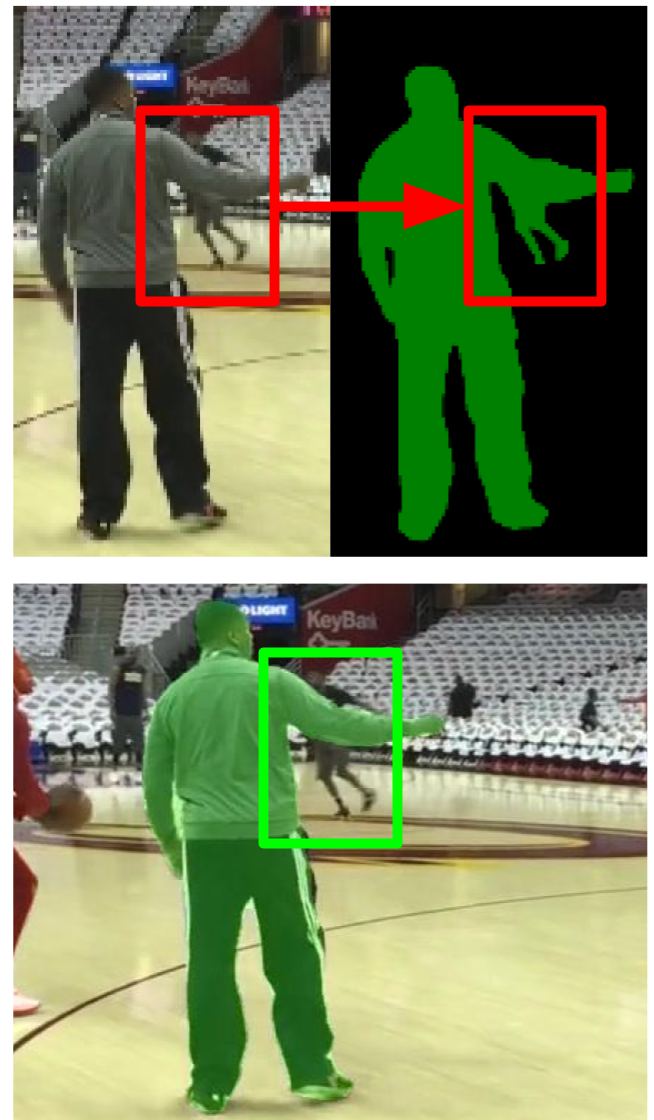}
        \caption{}
        \label{fig:mis1}
    \end{subfigure}
    \quad 
    \begin{subfigure}[b]{0.31\linewidth}
        \centering
        \includegraphics[height=6cm]{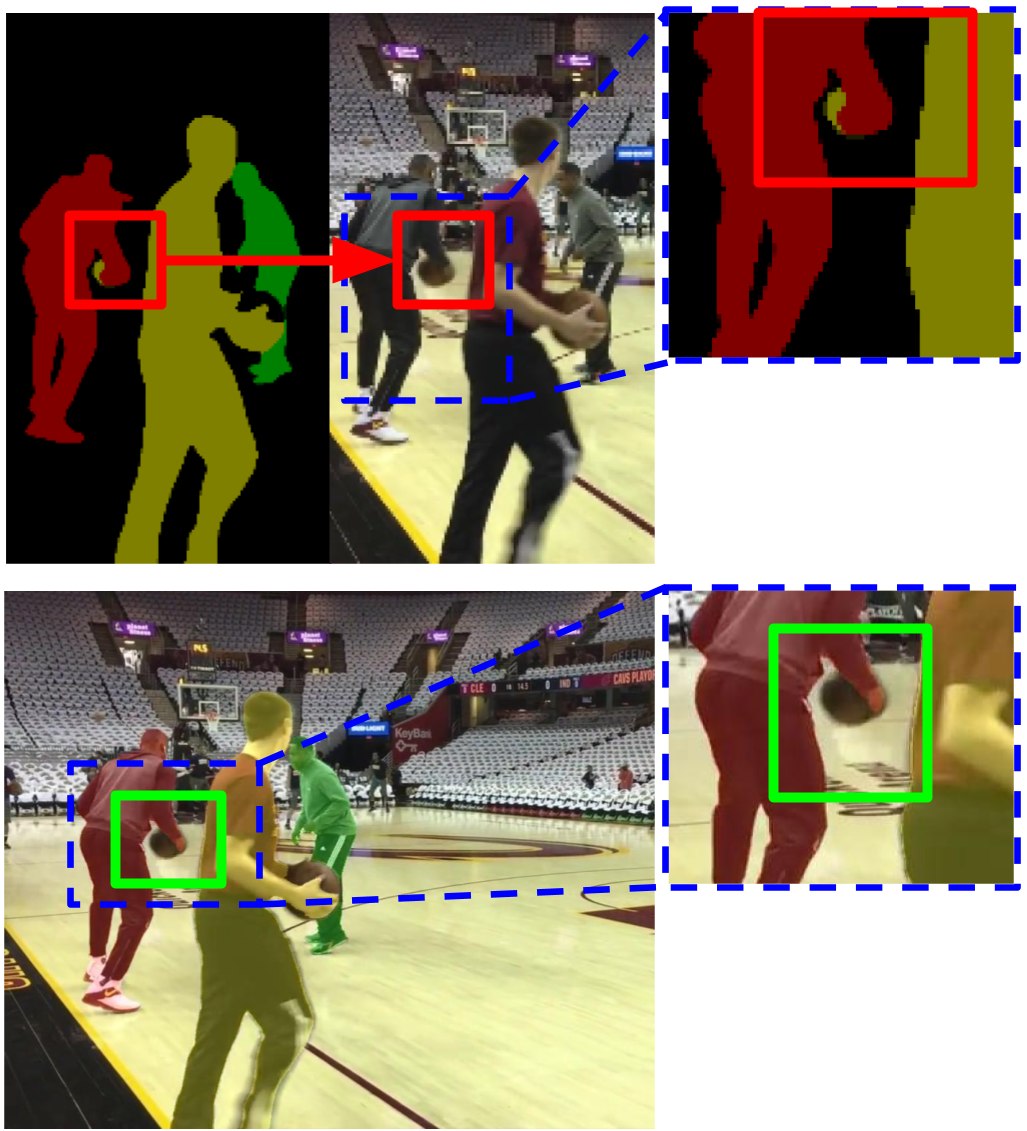}
        \caption{}
        \label{fig:mis2} 
    \end{subfigure}
    \quad
    \begin{subfigure}[b]{0.3\linewidth}
        \centering
        \includegraphics[height=6cm]{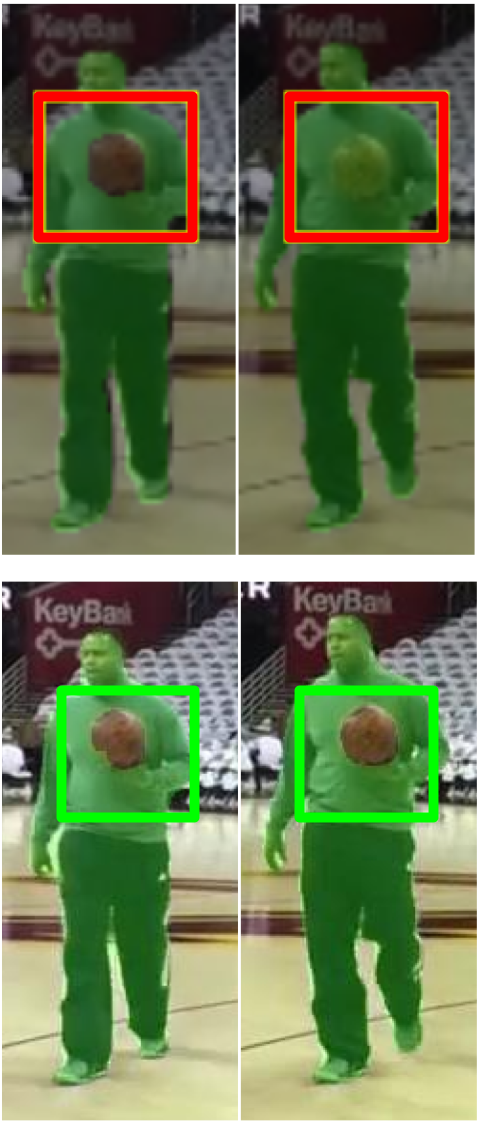}
        \caption{}
        \label{fig:mis3}
    \end{subfigure}
    \caption{Examples of ground-truth inconsistencies in LVOS (top) and the consistent annotation achieved with SAMannot (bottom): 
    (a) merging the masks of two distinct players 
    (b) inclusion of unlabeled objects, moreover, incorrectly given the same label as another instane 
    (c) temporal and structural inconsistency: the ball within the player's mask is labeled inconsistently across consecutive frames (alternating between allowing and preventing overlaps)
    }
    \label{fig:total_mis}
\end{figure}

During evaluation, we found several ground-truth annotation inconsistencies in LVOS, including merging masks from different instances, annotating some objects in a scene but omitting others, assigning labels to regions that do not belong to any instance (and are otherwise unlabeled), inconsistent handling of overlapping masks, and occasional holes in masks.
Figure~\ref{fig:total_mis} illustrates representative examples of these discrepancies.
These issues decrease the reported metric values, even though SAMannot’s annotations were semantically consistent in these cases.

\textbf{Known limitations}\newline
Despite its robustness, SAMannot inherits certain architectural limitations from the underlying SAM2 model regarding prompt density. 
When a single instance is over-prompted, specifically when a high number of point prompts (typically exceeding 8–10 points based on empirical observations) are applied relative to the instance's spatial extent, the model's attention mechanism may produce fragmented segmentation masks. 
This can manifest as internal 'holes' or the generation of non-contiguous, sparse regions instead of a single connected component. 
Users are therefore encouraged to utilize a minimal but strategic set of prompts, leveraging the model’s spatial priors rather than attempting exhaustive point coverage.

\section*{(2) Availability}
\vspace{0.5cm}
\section*{Operating system}

Linux and Windows. SAMannot has been tested on Ubuntu 22.04 LTS and Microsoft Windows 11. 

\section*{Programming language}
Python 3.10 

\section*{Additional system requirements}
Due to the memory addressing requirements of foundation models and the underlying deep learning frameworks, only 64-bit operating systems are supported.\newline

The system requires Python 3.10 or higher. \newline

While CPU-only inference is technically possible, a CUDA-enabled NVIDIA GPU with at least 6 GB of VRAM is strongly recommended for fast, interactive performance. 
Proper operation requires up-to-date NVIDIA drivers and the CUDA Toolkit (version 11.8 or later) to match the SAM2 inference engine's requirements.

\section*{Dependencies}
The core libraries of SAMannot include NumPy ($\ge$1.22.0), Pillow ($\ge$9.0.0), OpenCV-python ($\ge$4.7.0), Matplotlib ($\ge$3.5.0), and scikit-image ($\ge$0.21.0). 
The PyTorch and computer vision framework is built upon torch ($\ge$2.0.0), torchvision ($\ge$0.15.0), and timm ($\ge$0.6.12). 
Integration with the SAM2 model requires segment-anything, omegaconf ($\ge$2.3.0), supervision ($\ge$0.13.0), and pycocotools ($\ge$2.0.6). 
The graphical user interface and its styling are implemented using tk ($\ge$0.1.0) and ttkthemes. 
Additionally, the system utilizes hydra-core for configuration management and openpyxl for spreadsheet operations. 
To ensure robust performance on commodity hardware, system resource monitoring and GPU management are handled by psutil, pynvml, and nvidia-ml-py3. 

\section*{List of contributors}
\textbf{Author contributions:}

Gergely Dinya: Conceptualization, Software, Evaluation, Writing – Original Draft. 

András Gelencsér: Software Validation, Formal Analysis, Investigation, Evaluation, Writing – Original draft, Review \& Editing. 

Krisztina Kupán: Software Validation, Resources (Data Provision), Writing – Review. 

Clemens Küpper: Resources (Data Provision), Writing – Review. 

Kristóf Karacs: Writing – Original Draft, Review \& Editing. 

Anna Gelencsér-Horváth: Conceptualization, Methodology, Supervision, Evaluation, Writing – Original Draft, Review \& Editing.

\section*{Software location:}

{\bf Code repository}


\begin{itemize} 
	\item[]Name: GitHub
	\item[]Identifier: \url{https://samannot.github.io/}
	\item[]Licence: MIT License
	\item[]Date published: 2026-01-16
\end{itemize}

\section*{Language}
English

\section*{(3) Reuse potential}

The modular architecture and open-source nature of SAMannot provide significant potential for reuse and extension across diverse scientific domains. 

The software core functionality is domain-agnostic. 
Researchers in biomedicine can leverage the tool for segmenting cellular structures or tracking moving organelles in microscopy videos. 
In robotics and autonomous systems, the framework can be utilized to generate ground-truth data for rare edge cases in traffic or industrial environments where cloud-based labeling is prohibited due to confidentiality. 
Furthermore, the ability to handle long-term occlusions via SAM2 propagation makes it suitable for sports analytics and surveillance research.

The codebase is designed with a clear separation between the GUI, the SAM2 wrapper, and the data management modules, allowing for several paths of extension.
The specialized wrapper class is designed to decouple model-specific inference from the core application logic, allowing for the seamless integration of future foundation models, such as SAM3~\cite{carion2025sam3}, or alternative backends with minimal adjustments to the frontend interface. 
Furthermore, the framework's auto-prompting mechanism (presently utilizing topological skeletonization) can be readily adapted to incorporate domain-specific heuristics, such as utilizing DeepLabCut~\cite{dlc} keypoints as prior prompts to guide temporal propagation. 
Finally, the software architecture supports extensible, domain- and data-specific post-processing via custom scripts that can be integrated into the export pipeline. 
This allows users to automatically compute and export additional derived quantities beyond the previously described outputs.

We actively encourage community contributions and feedback. Potential contributors are invited to submit pull requests or open issues via the official GitHub repository. 
For long-term collaboration or specific integration queries, researchers may contact the corresponding author directly via email. 

Regarding technical support, the primary channel for bug reporting and feature requests is the GitHub Issue system, where we provide best-effort maintenance.
A step-by-step guide in the README and a short demo video are provided to lower the entry barrier for non-technical users.    

\begin{appendix}
\renewcommand\thesection{Appendix~\Alph{section}}

\section{Pop-up windows}
\label{sec:AppendixPopups}
\begin{figure}[H]
    \centering
    \includegraphics[width=0.7\linewidth]{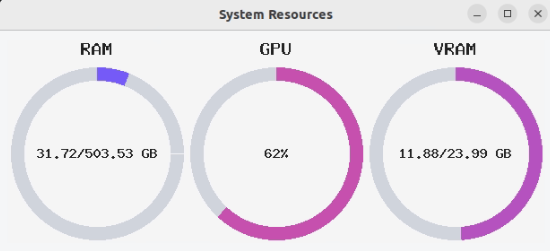}
    \caption{A system resources monitor, accessible via a pop-up from the main control window, provides real-time tracking of RAM usage, GPU utilization, and GPU VRAM occupancy.}
    \label{fig:popups}
\end{figure}

\begin{figure}[H]
    \centering
    \includegraphics[width=\linewidth]{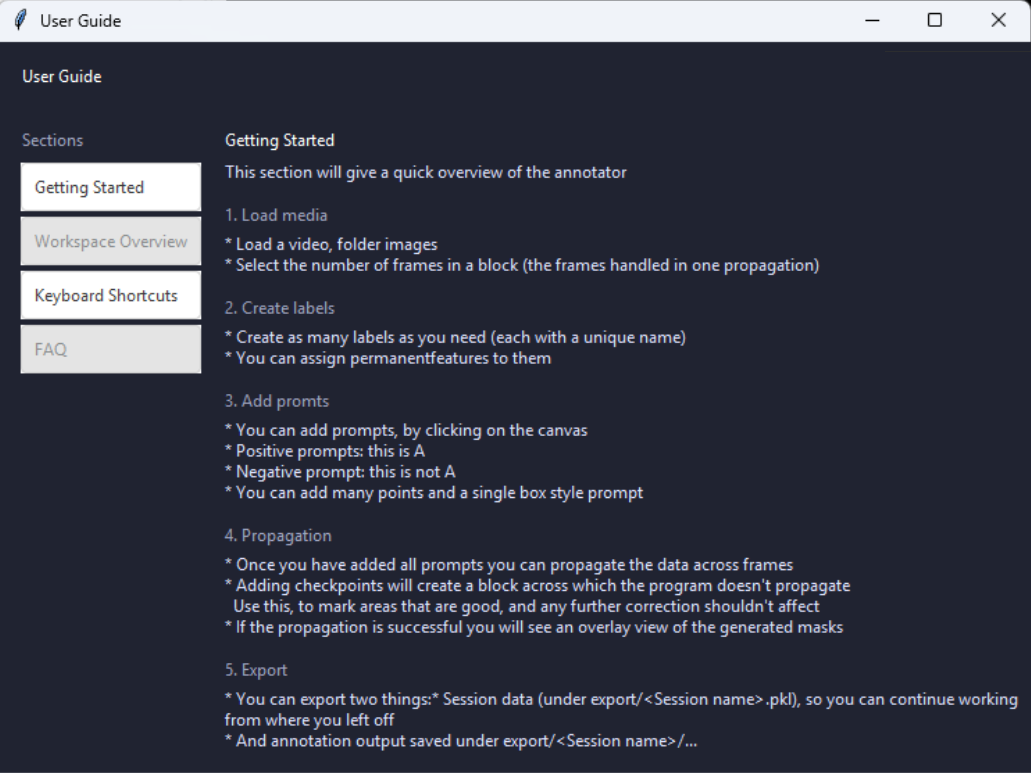}
    \caption{Illustration of User guide windows(A).}
    \label{fig:userguide1} 
\end{figure}

\begin{figure}[H]
    \centering
    \includegraphics[width=\linewidth]{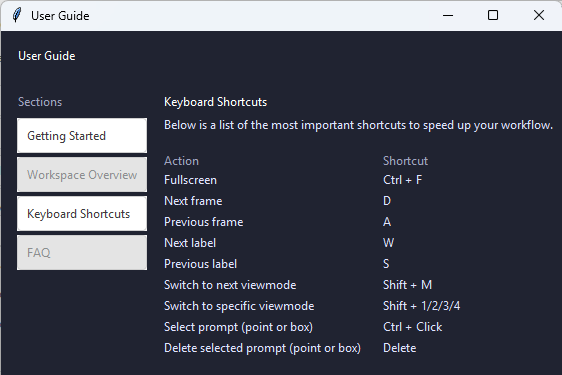}
    \caption{Illustration of User guide windows(B).}
    \label{fig:userguide2} 
\end{figure}

\section{Visualization of annotations}
\label{app:visAnnot}

\begin{figure}[H]
    \centering
    \begin{minipage}{0.82\linewidth}
        \centering
        \includegraphics[width=\linewidth]{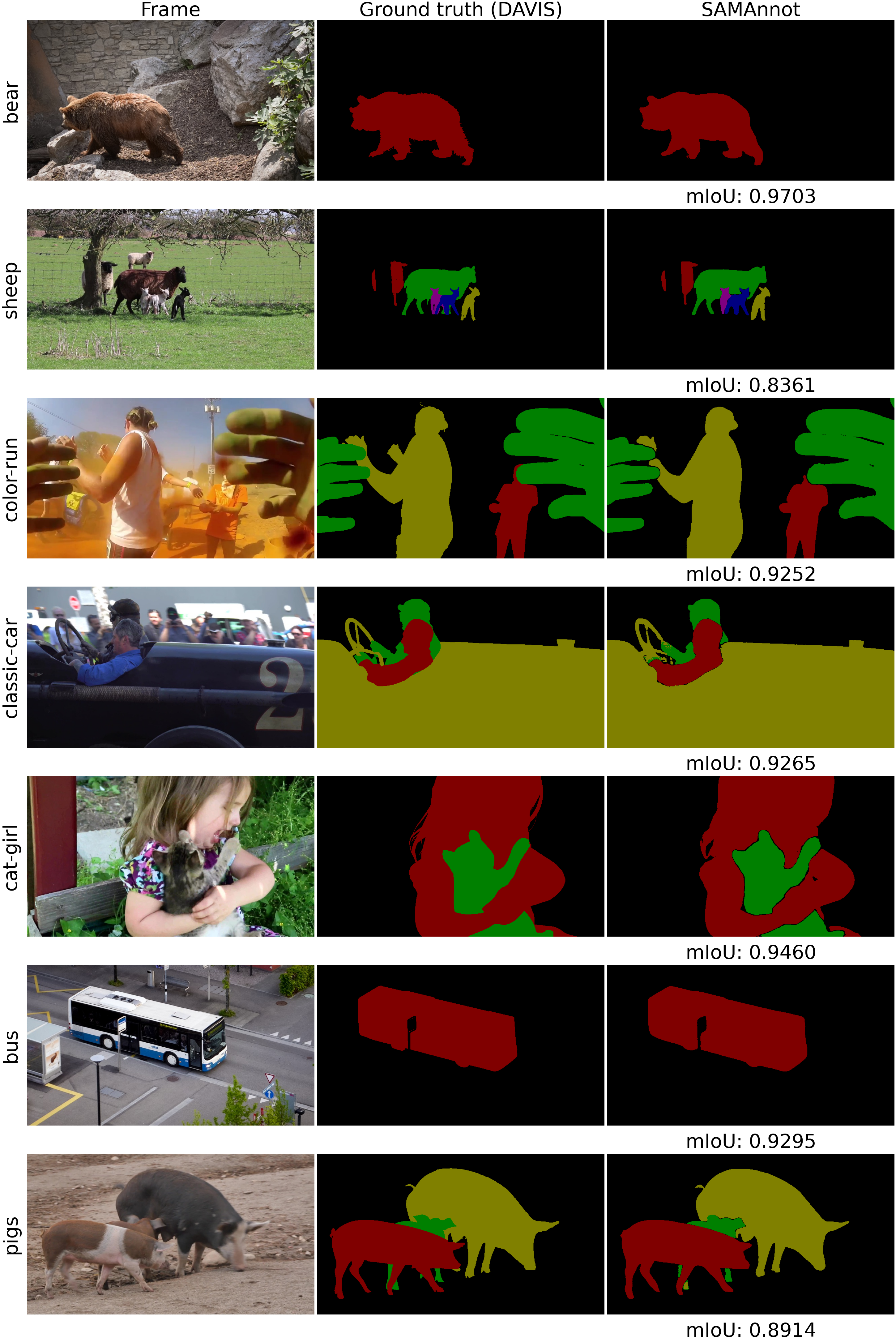}
        \caption{Qualitative examples of segmentation results on images from the DAVIS 2017 dataset~\cite{DAVIS}. The columns display the original video frame (left), the ground truth (middle), and the masks predicted by SAMannot (right).}
        \label{fig:totalgrid} 
    \end{minipage}
\end{figure}

\end{appendix}

\section*{Acknowledgements}
The authors would like to thank Viktor Karacs and Simon Vandepitte for alpha testing the application and providing feedback on the user experience during the early stages of development. We also thank Lili Stajer for developing the initial graphical user interface prototype. Finally, we thank Péter Zsoldos for technical contributions to SAM2 memory optimization and for analyzing relevant open-source issues. The authors are grateful to András Lőrincz for his useful suggestions and support.

\section*{Funding statement}
This work was supported by the funding of the Max Planck Society (to CK). 
This work was supported by project No. 2022-1.2.5-TÉT-IPARI-KR-2022-00015, funded by the National Research, Development and Innovation Fund, managed by the National Research, Development and Innovation Office, with the support of the Ministry of Culture and Innovation.

\section*{Competing interests}
The authors declare that they have no competing interests.

\medskip

\bibliographystyle{unsrt}
\bibliography{references}

\vspace{2cm}

\rule{\textwidth}{1pt}





\end{document}